\documentclass{article}

 \PassOptionsToPackage{square, numbers, compress,sort}{natbib}

\usepackage[final]{neurips_ts4h_2022}

\usepackage[utf8]{inputenc} %
\usepackage[T1]{fontenc}    %
\usepackage[backref=page]{hyperref}       %

\usepackage{url}            %
\usepackage{booktabs}       %
\usepackage{amsfonts}       %
\usepackage{nicefrac}       %
\usepackage{microtype}      %
\usepackage[table,svgnames]{xcolor}         %

\usepackage{graphicx}
\usepackage{caption}
\usepackage{subcaption}

\usepackage{amsmath,amsbsy,amsgen,amscd,amsfonts} %
\usepackage{cleveref}
\usepackage{bbm} %
\usepackage{mathrsfs} %
\usepackage{bbold}
\usepackage[utf8]{inputenc}
\usepackage{csquotes}

\providecommand{\mathbbm}{\mathbb} %

\newcommand{\1}{\mathbbm{1}} %

\newcommand{\R}{\mathbbm{R}}

\title{Learning Absorption Rates in Glucose-Insulin Dynamics from Meal Covariates}

\author{%
   Ke Alexander Wang\thanks{These authors contributed equally.} \\
   Stanford University\\
   \texttt{alxwang@cs.stanford.edu} \\
   \And
   Matthew E.~Levine\footnotemark[1] \\
   California Institute of Technology\\
   \texttt{mlevine@caltech.edu} \\
   \AND
   Jiaxin Shi\\
   Stanford University\\
   \texttt{jiaxins@stanford.edu} \\
   \And
   Emily B.~Fox \\
   Stanford University \& CZ Biohub\\
   \texttt{ebfox@stanford.edu}
}

\begin{document}

\maketitle

\begin{abstract}
Traditional models of glucose-insulin dynamics rely on heuristic parameterizations chosen to fit observations within a laboratory setting. However, these models cannot describe glucose dynamics in daily life. One source of failure is in their descriptions of glucose absorption rates after meal events. A meal's macronutritional content has nuanced effects on the absorption profile, which is difficult to model mechanistically. In this paper, we propose to learn the effects of macronutrition content from glucose-insulin data and meal covariates.  Given macronutrition information and meal times, we use a neural network to predict an individual's glucose absorption rate. We use this neural rate function as the control function in a differential equation of glucose dynamics, enabling end-to-end training. On simulated data, our approach is able to closely approximate true absorption rates, resulting in better forecast than heuristic parameterizations, despite only observing glucose, insulin, and macronutritional information.  Our work readily generalizes to meal events with higher-dimensional covariates, such as images, setting the stage for glucose dynamics models that are personalized to each individual's daily life. 
\end{abstract}

\section{Introduction}
\vspace{-2mm}
Type-1 diabetes is a chronic condition of glucose dysregulation that affects 9 million people around the world.
Decades of research have produced dozens of glucose-insulin dynamics models in order to understand the condition and help diabetics manage their daily lives.
These models are typically developed using physiological knowledge and validated in laboratory settings.
However, these mechanistic models are incomplete; they are not flexible enough to fit observations outside of controlled settings, due to unmodelled variables, unmodelled dynamics, and external influences. 
As a result, these mechanistic models fail to fully describe an individual's glycemic response to external inputs like nutrition.

Standard models, such as \citet{dallamanMealSimulationModel2007}, focus on the glycemic impact of carbohydrates in a meal---carbohydrates are broken down into glucose molecules, then absorbed into blood.
However, these models typically ignore other macronutrients, such as fat, fiber, and protein, which are known to contribute substantially to the amount and timing of glucose absorption into the blood.
Indeed, this phenomenon is the basis for the glycemic index of various foods.
In reality, individual glycemic responses to nutrition go beyond such a simple characterization.
For example, \citet{zeeviPersonalizedNutritionPrediction2015} identified multiple patient sub-groups with different glycemic responses to complex foods.

In our paper, we propose a method that can leverage real-world nutrition and glucose-insulin measurements to improve the fidelity of existing mechanistic models.
While we tailor this approach to the specific application of type-1 diabetes, we note that our methodology fits within a broad paradigm of hybrid modeling of dynamical systems \citep{willardIntegratingScientificKnowledge2022,levineFrameworkMachineLearning2021,Miller2021BreimanTwoCultures,rico-martinezContinuoustimeNonlinearSignal1994}.
These approaches can improve mechanistic ODEs using flexible components that learn from observations of the system and its external controls.

\section{Background on modelling glucose-insulin dynamics}
\vspace{-2mm}
Our paper builds on the tradition of modelling physiological dynamics via ordinary differential equations (ODEs), 
\citep{bergmanQuantitativeEstimationInsulin1979,sturis91,dallamanMealSimulationModel2007,hovorka08,mariMathematicalModelingPhysiological2020}.
Traditional models consider ODEs of the form $\dot{x}(t) = f(t, x(t)) + u(t)$,
where $x \in \R^n$ denotes physiologic states, $f: \R^n \to \R^n$ encodes mechanistic knowledge of their interactions, 
and $u: \R \to \R^n$ represents external time-varying inputs into the system.
Significant effort has gone towards identifying $u$ from insulin, exercise, and meal data, but $u$ is typically represented via a gastrointestinal ODE model \citep{elashoffAnalysisGastricEmptying1982,dallamanSystemModelOral2006} or via hand-chosen functional forms
\citep{hovorkaNonlinearModelPredictive2004,herreroSimpleRobustMethod2012,liuEnhancingBloodGlucose}. Both approaches for representing meals depend only on carbohydrate consumption and do not consider other macronutrient quantities.

Our paper considers the minimal model of glucose-insulin dynamics by \citet{bergmanQuantitativeEstimationInsulin1979}:
\begin{subequations}
   \label{eq:bergman}
  \begin{align}
  \dot{G}(t) &= -c_1 [G(t) - G_b] - G(t)X(t) + u_G(t) \\
  \dot{X}(t) &= -c_2 X(t) + c_3[I(t) - I_b]\\
  \dot{I}(t) &= -c_4 [I(t) - I_b] + u_I(t)
    \end{align}
\end{subequations}
where $x=(G,X,I)$ and $u=(u_G, u_I)$.
Here, $G: \R\to \R$ represents plasma glucose concentration,
$I: \R\to\R$ represents plasma insulin concentration,
$X: \R\to\R$ represents the effect of insulin on glucose,
$G_b, I_b \in \R$ represent basal glucose and insulin levels, respectively, and $c_1, c_2, c_3, c_4 \in \R$ represent rate constants for the interactions.
Importantly, $u_G:\R\to \R$ represents the appearance of glucose in the blood (e.g. absorbed from nutrition in the gut) and $u_I: \R \to \R$ represents the appearance of insulin in the blood (e.g. absorbed from subcutaneous injection or drip).
See \citet{Gallardo-Hernandez2022MinimallyInvasiveEfficientMethod} for a modern exposition and the units of each quantity.

\paragraph{Modelling nutrition absorption from discrete meal events.}
When simulating the daily management of diabetes, the \emph{continuous} functions $u_G, u_I$ are typically derived from observed \emph{discrete} events (e.g. meals and insulin injections).
Each discrete-time event $e_i=(t_i, m_i)$ consists of a timestamp $t_i$ and a covariate $m_i$.
If $e_i$ is a meal event, $m_i$ may consist of macronutritional information, an image of the food, or both.
Pharmacodynamics models are often used to map the insulin dose to a continuous absorption profile $u_I$ that is compatible with the above model.
However, the dependence of glucose absorption $u_G$ on full macronutritional content of a meal event is less well-understood; thus \emph{we focus on modelling $u_G$ in this paper}.

Mechanistic $u_G$ models often derive $u_G$ as the solution to another set of heuristic ODEs\citep{dallamanMealSimulationModel2007}.
However, this approach introduces additional handcrafted parameterizations to explain quantities that are unobservable outside of the lab setting, such as the glucose concentration in the stomach over time after a meal. 
A simpler yet effective approach is to directly model $u_G$ phenomenologically, and estimate it from data \citep{herreroSimpleRobustMethod2012,liuEnhancingBloodGlucose}.
Instead of deriving $u_G$ from an intricate model of the human body, this approach represents $u_G$ directly using a parametric function adapted from data.

\section{Phenomenologically modelling the absorption rate}\label{sec:phenomenological-models}
\vspace{-2mm}
Let each meal event $i$ be $e_i=(t_i, m_i)$ where $t_i\in \R$ is the meal time and $m_i \in \R^M$ is a vector of meal covariates, such as its macronutrition content or even a photo of the food.
We assume we have data on a set $E$ of these meal events.
For each meal $i$, we associate a parametric function $a_i: \R_+ \to \R_+$, such that $a_i(t)$ is the absorption rate of the meal at time $t$.
The overall control function $u_G$ is then a sum over the events:
\begin{equation}
    \label{eq:general-control}
    u_G(t) = \sum_{i=1}^{|E|} a_i(t).%
\end{equation}
$a_i$ is usually compactly supported, since meals only affects glucose locally in time.
Decomposing $u_G$ into a sum allows us to model the effect of each meal individually, instead of all at once.

A simple heuristic choice is a square function 
$a_i(t) = g_i \1_{[0, w)}(t - t_i) / w$ where $w$ is the width of the square as a free parameter and $g_i \in \R$ is the amount of glucose produced from the meal.
Another choice is the bump function $a_i(t) = g_i \1_{[0, \infty)}(t-t_i)(e^{-b_1(t-t_i)} - e^{-b_2(t-t_i)})/b_3$ where $b_1$ and $b_2$ are free parameters and $b_3$ is a normalization constant \citep{albersPersonalizedGlucoseForecasting2017,albersSimpleModelingFramework2020}.
For both choices, $g_i$ must be estimated by the patient or by a nutritionist (e.g. when $m_i$ is a food image), which can be highly inaccurate.
More importantly, the \emph{shape} of these parameterizations does not depend on  $m_i$, even though foods vary in absorption profiles.

\paragraph{A neural phenomenological model.}
The form of Equation~\eqref{eq:general-control} suggests a natural extension that takes advantage of the flexiblity of neural networks.
Given a meal event $e_i=(t_i, m_i)$, we model its absorption rate using a neural network $a_\theta$ such that
\begin{equation}
    \label{eq:nn-control}
    a_i(t) = g_i \cdot a_\theta(t - t_i, m_i) \1_{[0, \infty)}(t-t_i).
\end{equation}
We make use of the estimated glucose content $g_i$ following prior approaches since it is often already available in the meals dataset, and gives an expert-informed glucose absorption scale factor.
Alternatively, $g_i$ can be included as another input to $a_\theta$ instead of being a multiplicative constant.
Even if the estimated $g_i$ is inaccurate, $a_\theta$ has the flexiblity to rescale $g_i$ based on the observed $m_i$.
Most importantly, our parameterization differs in that its \emph{shape} can adapt to the meal covariates $m_i$.
We share one neural network $a_\theta$ across all meal events, allowing it to generalize to macronutritional information similar to, but not exactly the same as, meals from the training set.
Altogether, Equations~\eqref{eq:bergman},\eqref{eq:general-control},\eqref{eq:nn-control} define our neural differential equation model. 

\paragraph{End-to-end training on partial observations.}
Having defined our parametric function, we now discuss how to learn the parameters $\theta$ in a setting that is realistic to settings outside of the laboratory.
Recent technologies like continuous glucose monitors and artificial pancreases enable real-time measurements of glucose levels and insulin dosage.
However, most of a patient's physiological state is unobserved.
Within Equation~\eqref{eq:bergman}, we do not observe insulin $I$ and its effect $X$.

Let $x$ be the state of our differential equation from Equation~\eqref{eq:bergman}.
We assume our temporal data consists of noisy partial observations over time $\{(t_k, y_k)\}_{k=1}^T$, where $y_k = Hx(t_k) + \varepsilon$.
We assume the projection operator $H: (G,X,I) \mapsto (G, 0, 0)$ and $\varepsilon$ is a zero-mean i.i.d. noise process.
Given initial condition $x(t_0) = x_0$, we can numerically integrate Equation~\eqref{eq:bergman} with a given $u_I$ and our parameterized $u_G(\cdot\ ; \theta)$ to obtain an estimate $\hat y(t_k) = H\hat x(t_k)$ where $\hat x(t_k) = \text{Integrate}(f, u, x_0, t_0, t_k)$.
We then minimize the mean squared error objective $L(\theta) = \sum_{k=1}^T \|\hat y(t_k) - y_k\|_2^2/T$ with respect to $\theta$ to fit our parametric model \citep{Finzi2020SimplifyingHamiltonianLagrangian}.
However, this procedure requires us to know $x_0$, which is not fully observed in practice.

Many methods exist for performing such under-determined state and parameter estimation; often, the state-estimation component is performed using filtering or smoothing \citep{brajardCombiningDataAssimilation2020,levineFrameworkMachineLearning2021,chenAutodifferentiableEnsembleKalman2021,chenNeuralOrdinaryDifferential2018,rubanovaLatentOrdinaryDifferential2019}, but can also be learnt through other data-driven \citep{ayedLearningDynamicalSystems2019,kemethInitializingLSTMInternal2021} or gradient-descent \citep{ouala_learning_2020} methods.
In our experiments, we estimate an initial state $x_0$ by using a sequence of $F$ observations $(G(t_{-F+1}), G(t_{-F+2}), \ldots, G(t_0))$ as a forcing function when forward integrating Equation~\eqref{eq:bergman}, described in Section 4.3 of  \citet{levineFrameworkMachineLearning2021}.
This simple procedure was sufficient for our model to learn a good $\theta$, likely due to the rapidly decaying autocorrelation of \eqref{eq:bergman}.

\section{Experiments}
\vspace{-2mm}
We evaluate our proposed method on simulated data.
We simulate 28 days worth of glucose, insulin, and meal data for one virtual patient using Equation~\eqref{eq:bergman}.
We evaluate our method against baseline methods with and without glucose observation noise.
We also evaluate each method in the realistic setting where the \emph{time} of each meal is noisily reported, since in daily life, the recorded meal time is often only approximately correct.

\begin{figure}
     \captionsetup[subfigure]{labelformat=empty}
     \centering
     \begin{subfigure}[b]{0.49\textwidth}
         \centering
        \includegraphics[width=\linewidth]{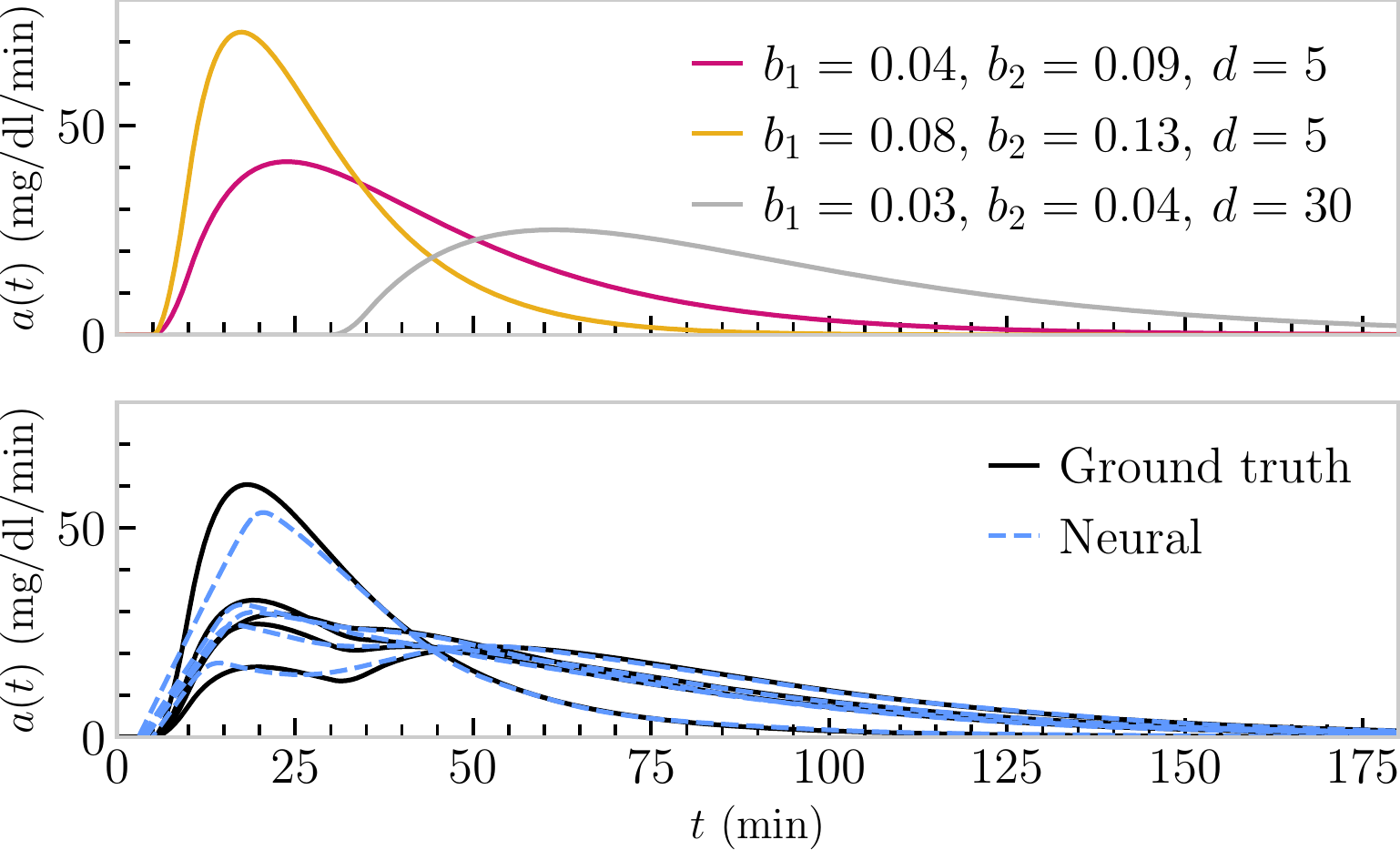}
     \end{subfigure}
     \hfill
     \begin{subfigure}[b]{0.49\textwidth}
         \centering
            \includegraphics[width=\linewidth]{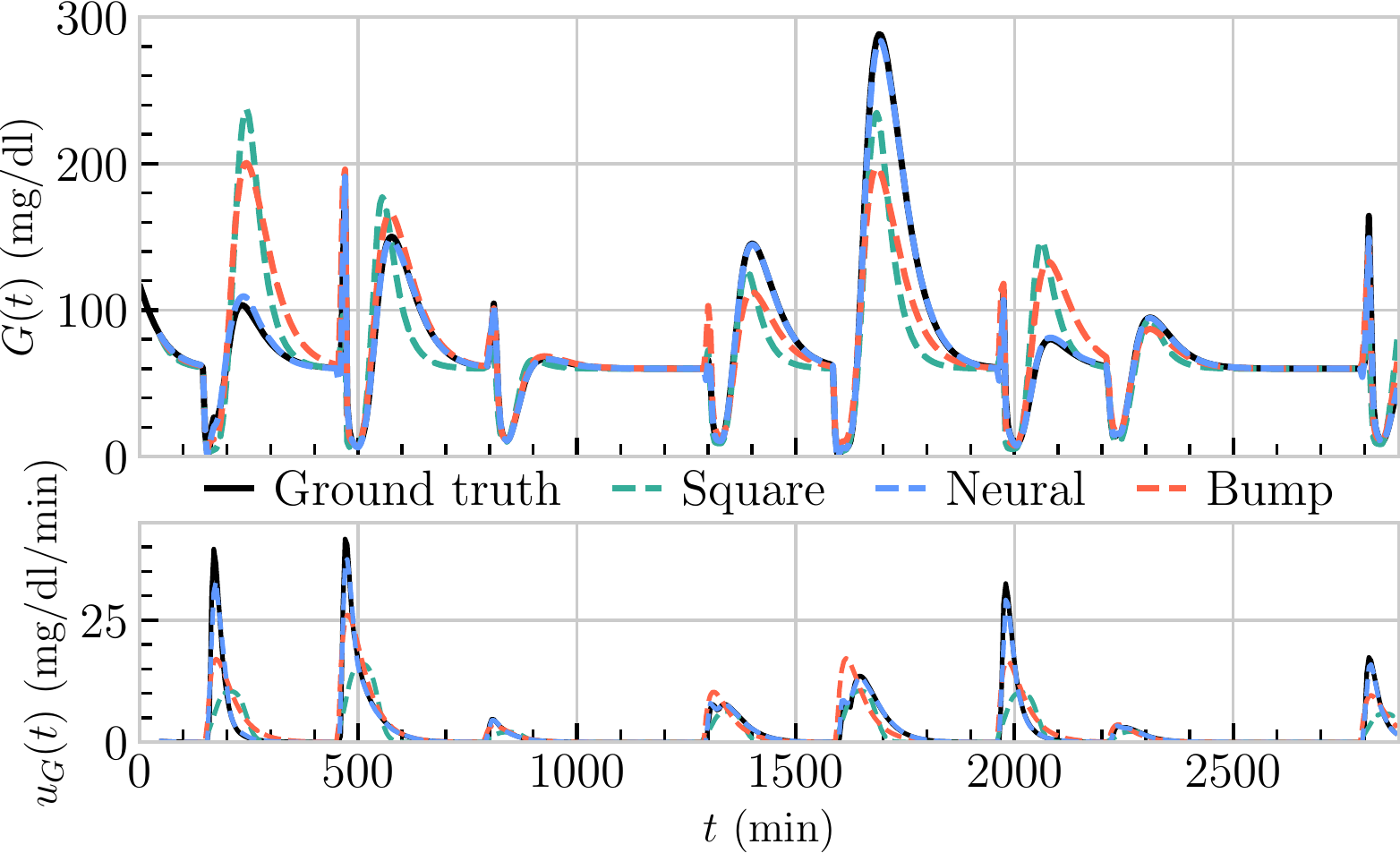}
     \end{subfigure}
     \caption{\emph{Left}: (Top) ``Absorption templates'' used to generate $u_G$. (Bottom) 5 Samples of ground truth and learned $u_G$ for meals from the test set. \emph{Right}: Glucose forecast and predicted absorption rates of each model over a 2 day window from the test trajectory.}
     \label{fig:absorption_and_forecast}
     \vspace{-6mm}
\end{figure}

\paragraph{Data generation.} For each day, we generate four meals: breakfast, lunch, dinner, and a late snack. Meals occur uniformly at random within 6-9AM, 11AM-2:30PM, 5-8PM, and 10-11PM, respectively.
Each meal contains a glucose amount uniformly random within 5-65g, 20-70g, 40-100g, and 5-15g respectively.
For each meal event $i$, we convert grams of glucose to plasma glucose concentration, assuming the individual has 50dl of blood, and use the result as $g_i$.
To simulate different absorption profiles, each meal is a convex mixture of three ``absorption templates''.
Each template $j$ is given by delayed bump function $a^j(t) \propto g_i \1_{[0, \infty)}(t-t_i-d)(e^{-b_1(t-t_i-d)} - e^{-b_2(t-t_i-d)})$, each with its own set of parameters $(b_1, b_2, d) \in \{(0.04, 0.09,5\text{min}), (0.08, 0.13, 5\text{min}), (0.03, 0.04, 30\text{min})\}$, visualized in Figure~\ref{fig:absorption_and_forecast}.
The templates represent regular absorption, fast absorption, and slow absorption, respectively.
The macronutrition of meal $i$ is then the vector of mixture coefficients $m_i \in \R^3$ such that meal $i$ has absorption profile $a_i(t) =\sum_{j=1}^3 m_{ij} a^j(t)$.
To ensure $a_i$ is smooth, we average each value $a_i(t)$ with a grid of 50 points from the past 5 minutes.

For each meal time $t_i$, we simulate an insulin bolus dose at a time sampled from $\mathcal{N}(t_i, (10\text{min})^2)$.
We sample a glucose to insulin conversion %
for each meal from $\mathcal{N}(7\text{g/U}, (1\text{g/U})^2)$.
To simulate imperfect measurements, we add a relative $N(0, 0.05^2)$ observation noise.
To simulate imperfect meal time recordings, we add $N(5\text{min}, (2.5\text{min})^2)$ noise to meal times.
We use a square function $u_I$, corresponding to a constant insulin absorption rate, over $30$ minutes, which we assume to be known to every model.
We use parameters from \citet{andersenBayesianApproachBergman2003} for Equation~\ref{eq:bergman}, and we use Euler integration with a step size of $0.1$ minutes to produce an observation every $5$ minutes.

\paragraph{Experimental setup.}
We split our generated data temporally into 3 disjoint training, validation, and testing trajectories.
We optimize using Adam \citep{adam} for 1000 iterations, with a half-period cosine learning rate schedule following a linear ramp up to $0.2$ over the first 30 iterations.
We use minibatches of 512 sequences of 4 hour windows (48 observations) and use 10 observations for estimating the initial condition.
We minimize the mean squared error on the observed glucose values with respect to the parameters $\theta$ of $a_\theta$, keeping the other parameters of Equation~\eqref{eq:bergman} fixed.
We parameterize our neural $a_\theta$ using a feedforward network with 2 hidden layers of 64 units and GELU activations.
We found that appropriately scaling the input and outputs of $a_i$ is crucial for stable optimization.

\begin{table}[h!]
\centering
\newcommand{\cellhi}{\cellcolor{RoyalBlue!15}}
\begin{tabular}{ccccc}
    \toprule
    \phantom{} & \multicolumn{2}{c}{Exact timestamps} & \multicolumn{2}{c}{Noisy timestamps} \\
    \cmidrule(lr){2-5}
    $a_i$ & Exact observations & Noisy observations & Exact observations & Noisy observations\\
    \cmidrule(lr){1-1} \cmidrule(lr){2-3} \cmidrule(lr){4-5}
    Neural & \cellhi 0.95{\tiny mg/dl} & \cellhi 3.66{\tiny mg/dl} & \cellhi 1.48{\tiny mg/dl} & \cellhi 3.63{\tiny mg/dl} \\
    Bump   & 9.52{\tiny mg/dl} & 10.11{\tiny mg/dl} & 9.53{\tiny mg/dl} & 10.24{\tiny mg/dl} \\
    Square & 11.60{\tiny mg/dl} & 11.53{\tiny mg/dl} & 11.65{\tiny mg/dl} & 11.56{\tiny mg/dl} \\
    \bottomrule
\end{tabular}
\vspace{1mm}
\caption{Forecast RMSE computed over all possible 4 hour windows of the test set trajectory, reflecting the window size used for training.}
\label{tab:rmse}
\vspace{-6mm}
\end{table}

\paragraph{Evaluations.}
We compare our neural absorption function against the two common parameterizations of $u_G$ from Section~\ref{sec:phenomenological-models}, fit via gradient-based optimization.
We approximate the piece-wise constant square function using a difference of sigmoids; otherwise the width cannot be learned.
Our neural model is able to closely approximate the ground truth $u_G$, especially in the tails, as shown in Figure~\ref{fig:absorption_and_forecast} (left).
This results in significantly better forecasts, and our neural model closely tracks the ground truth glucose values and absorption rates, \emph{even extrapolating to durations much longer than what was seen in training}. 
We visualize such long term forecasts in Figure~\ref{fig:absorption_and_forecast} (right).
We also report the forecast RMSE on the test set in Table~\ref{tab:rmse}.
Our neural model attains lower forecast errors across all settings.
In the noiseless case, our neural model is 10x more accurate than heuristic parameterizations.
The RMSEs generally increase as we add noise, though the bump and square functions are already such poor forecasters that noise does not worsen their errors significantly.

\section{Discussion}
\vspace{-2mm}
Our experiments show that our proposed method is a promising way to learn absorption profiles that depend on macronutritional information.
Our approach readily generalizes to handle arbitrary meal covariates beyond macronutritional information, such as food images or descriptions.
Although this paper only uses synthetic data, our method can complement any glucose dynamics model of real-world data.
Learning accurate dynamics from data, however, remains a challenging problem.
We see our method as a vital component in future data-driven hybrid models of glucose-insulin dynamics.

\begin{ack}
This work was supported in part by AFOSR Grant FA9550-21-1-0397, ONR Grant N00014-22-1-2110, and the Stanford Institute for Human-Centered Artificial Intelligence
(HAI). KAW was partially supported by Stanford Data Science as a Data Science Scholar.
MEL was supported by the National Science Foundation Graduate Research Fellowship under Grant No. DGE‐1745301.  EBF is a Chan Zuckerberg Biohub investigator.

\end{ack}

\bibliographystyle{plainnat}
\bibliography{Dynamode}

\end{document}